\pdfoutput=1
\documentclass[letterpaper]{article} 
\usepackage[]{aaai23}  
\usepackage{times}  
\usepackage{helvet}  
\usepackage{courier}  
\usepackage[hyphens]{url}  
\usepackage{graphicx} 
\usepackage{natbib}  
\usepackage{caption} 
\usepackage{algorithm}
\usepackage{algorithmic}
\usepackage{diagbox}
\usepackage{graphicx}
\usepackage{multirow}
\usepackage{color}
\usepackage{amssymb}
\usepackage{hyperref}

\newcommand{\cut}[1]{}

\usepackage{amsmath}

\usepackage{newfloat}
\usepackage{listings}
\DeclareCaptionStyle{ruled}{labelfont=normalfont,labelsep=colon,strut=off} 
\floatstyle{ruled}
\newfloat{listing}{tb}{lst}{}
\floatname{listing}{Listing}
\usepackage{bibentry}

\begin{document}
%

\title{Dynamic Memory-based Curiosity: A Bootstrap Approach for Exploration}

\author{
	Zijian Gao\textsuperscript{\rm 1}, YiYing Li\textsuperscript{\rm 2}, Kele Xu\textsuperscript{\rm 1}\thanks{Corresponding author: Kele Xu}, Yuanzhao Zhai\textsuperscript{\rm 1},\\ Dawei Feng\textsuperscript{\rm 1}, Bo Ding\textsuperscript{\rm 1}, XinJun Mao\textsuperscript{\rm 1}, Huaimin Wang\textsuperscript{\rm 1} 
}
\affiliations{
	\textsuperscript{\rm 1} National University of Defense Technology, Changsha, China\\
	\textsuperscript{\rm 2} Artificial Intelligence Research Center, DII, Beijing, China\\
	
	kelele.xu@gmail.com	\\				
}
\maketitle
\begin{abstract}
\begin{quote}
The sparsity of extrinsic rewards poses a serious challenge for reinforcement learning (RL). Currently, many efforts have been made on curiosity which can provide a representative intrinsic reward for effective exploration. However, the challenge is still far from being solved. In this paper, we present a novel curiosity for RL, named DyMeCu, which stands for Dynamic Memory-based Curiosity. Inspired by human curiosity and information theory, DyMeCu consists of a dynamic memory and dual online learners. The curiosity arouses if memorized information can not deal with the current state, and the information gap between dual learners can be formulated as the intrinsic reward for agents, and then such state information can be consolidated into the dynamic memory. Compared with previous curiosity methods, DyMeCu can better mimic human curiosity with dynamic memory, and the memory module can be dynamically grown based on a bootstrap paradigm with dual learners. On multiple benchmarks including DeepMind Control Suite and Atari Suite, large-scale empirical experiments are conducted and the results demonstrate that DyMeCu outperforms competitive curiosity-based methods with or without extrinsic rewards. We will release the code to enhance reproducibility.\footnote{© 2023 IEEE.  Personal use of this material is permitted.  Permission from IEEE must be obtained for all other uses, in any current or future media, including reprinting/republishing this material for advertising or promotional purposes, creating new collective works, for resale or redistribution to servers or lists, or reuse of any copyrighted component of this work in other works}
\end{quote}
\end{abstract}

\noindent 

\section{Introduction}
\begin{figure*}[t] 
    \centering
    \includegraphics[width=0.85\textwidth]{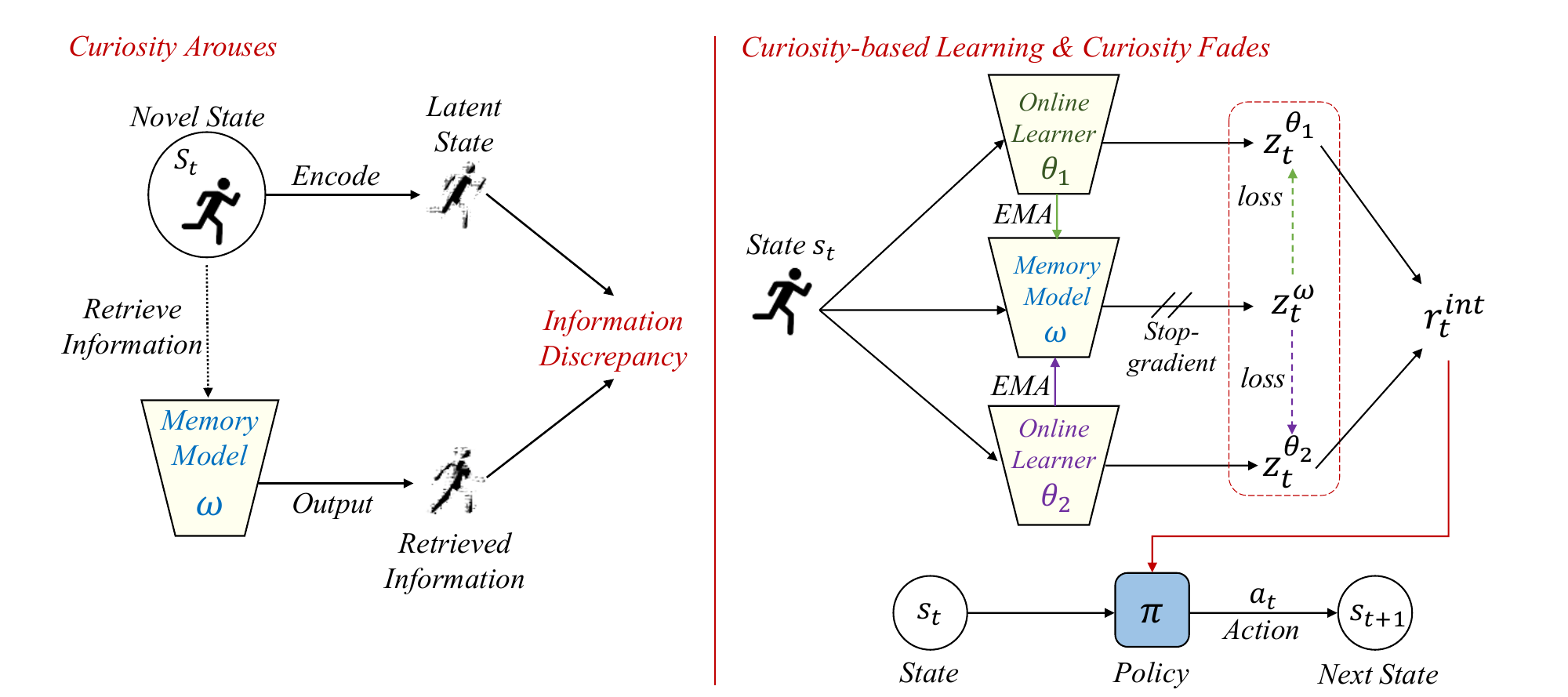}   
    \caption{
    DyMeCu employs a novel learning framework to build the intrinsic reward for RL, which consists of a dynamic memory and dual online learners. 
    The information discrepancy of the current state compared with the retrieved information from the memory makes curiosity arouse. We get the curiosity-based intrinsic reward for agent learning by calculating the information gap between dual online learners. Then the state information can be dynamically consolidated into the memory in the bootstrap paradigm for curiosity fading.} 
    \label{method}
\end{figure*}

Despite the success of reinforcement learning (RL) on sequential decision-making tasks~\cite{atari,game2,game3}, many current methods struggle with sparse extrinsic rewards.
To cope with the sparsity, curiosity provides a representative intrinsic reward that can encourage agents to explore new states. Designing algorithms to efficiently construct curiosity can be a key component in RL systems. %
Previous research has shown that intrinsic rewards can help alleviate the issues resulting from the lacking of dense extrinsic rewards~\cite{liu2021behavior,tao2020novelty,yang2021exploration}. 

For human learning, curiosity motivates people to seek and retain more information through exploration in the environment ~\cite{burda2018large,ryan2000intrinsic,smith2005development}.
The process of arousing and satisfying curiosity can be summed up as one cycle: when a person encounters a problem, he/she will first try to solve it by retrieving information from memory. If retrieval from memory fails, he/she realizes that the current memorized information is insufficient solve the problem. A conscious awareness of information discrepancy then sparks curiosity about the problem, and curiosity stimulates the search for new information. Once the information discrepancy is eliminated, people may have no further curiosity to learn more about the current problem until another problem is encountered~\cite{rotgans2017role,silvia2017curiosity}. Human curiosity is constantly consolidated based on the dynamic memory, which consists of the encoding, storing, and retrieving information stage~\cite{replay}. As the curiosity fades, additional information is consolidated into the memory. The consolidation results in the forming of new dynamic memories, which depends on the hippocampus~\cite{hip}.


Many attempts have been made to build curiosity in RL, which fall into two main categories: count-based and prediction-based. However, such curiosity is very different from human curiosity, and the problem is far from solved. Taking the Random Network Distillation (RND)~\cite{RND} method as an example, RND initializes a random fixed target network with state embeddings, and trains another prediction network to fit the output of the target network. A random fixed target network can be regarded as a random fixed memory, so that RND cannot retain contextual knowledge about the environment~\cite{yang2021exploration}.
Without dynamically incorporating contextual information into memory, random features may not be sufficient to interpret dynamic environments. Therefore, this kind of curiosity is evaluated in a non-developmental way, which severely limits the performance of curiosity in RL.



\footnotetext[1]{The term \emph{bootstrap} is used in this text in its colloquial meaning rather than its statistical connotation.}

In this work, to mimic human curiosity, we formalize and investigate a Dynamic Memory-based Curiosity mechanism, named DyMeCu. Inspired by the bootstrap paradigm~\cite{bootstrap1,bootstrap2,bootstrap3}, we construct dual online learners to learn the latent state to formulate dynamic memory model (Figure~\ref{method}).
On the one hand, state information can be consolidated to the memory via the exponential moving average (EMA) ~\cite{haynes2012exponential,klinker2011exponential,grebenkov2014following} of dual learners' parameters. The bootstrap paradigm, on the other hand, utilizes supervised signals from memory to improve dual learners' encoding ability.
Furthermore, the curiosity is measured by the information gap between the dual learners, which is essentially an uncertainty estimation of given state based on dynamic memory ~\cite{mai2022sample,liu2020simple,abdar2021review}.

\cut{On the one hand, the paradigm not only enables learners to consolidate contextual state information into memory but also to learn from memory, which is referred as the stable target network and whose parameters are an exponential moving average (EMA) of the learners' parameters ~\cite{haynes2012exponential,klinker2011exponential,grebenkov2014following}.
On the other hand, the proposed dual learners can be utilized to assess curiosity. The less the information is consolidated in memory and the greater the gap between learners, the greater the curiosity can be.
From another perspective, using the information gap between dual learners to assess curiosity is essentially an uncertainty quantification of state information in memory ~\cite{mai2022sample,liu2020simple,gal2016dropout,abdar2021review}.}

In brief, our contribution in this paper is:

\begin{itemize}

    \item We firstly analyze the shortcomings of previous curiosity-based intrinsic reward methods\cut{from the perspective of memory and information theory}, and suggest to mimic human curiosity leveraging a dynamic memory instead of a fixed one, based on the information theory.

    \item We propose a novel and practicable intrinsic reward method for RL agents, named DyMeCu (Dynamic Memory-based Curiosity), which consists of a dynamic memory and dual online learners, and thus can measure the curiosity and consolidate the information in a feasible way. Meanwhile, different strategies are explored to further improve the performance of DyMeCu.
    
    \item On multiple benchmarks including DeepMind Control Suite (DMC) \cite{dm_control} and Atari Suite \cite{atari}, large-scale empirical experiments demonstrate that DyMeCu outperforms the other competitive curiosity-based methods and pre-training strategies. 

\end{itemize}

\section{Related Work}
\label{relatedwork}
\subsection{Curiosity-Based Intrinsic Reward}
In RL, the exploration issue is a long standing challenge. Previous attempts suggest that: if there is no additional reward, exploration can be regarded as a hunt for information theoretically, which also can be viewed as the curiosity~\cite{berlyne1950novelty,schmidhuber1991possibility,kidd2015psychology,de2018curiosity,jaegle2019visual,friston2016active,peterson2021curiosity}.
One intuitive formulation of curiosity is the count-based methods, where the less visited state has more state novelty for exploration. But it can not scale to large-scale or continuous state spaces~\cite{kearns2002near,charikar2002similarity}. 
Inspired by count-based methods, RND calculates the state novelty by distilling a random fixed network (target network) into another prediction network (predictor network). The predictor network is trained to minimize the prediction error for each state and take the prediction error as the intrinsic reward.
Apart from count-based methods, prediction-based methods also show competitive or better performance by modeling the environment dynamics~\cite{ICM,Disagreemet,active,burda2018large}. With the assumption that more visited state-action pairs will result in more accurate prediction, the intrinsic reward can be applied as the variance of predictions of ensembles or the distance between prediction states and true states, such as the Disagreement method~\cite{Disagreemet} and ICM~\cite{ICM} method. 
There have been few attempts to design a curiosity that contains memory and effectively uses information consolidated in memory, which however is the main goal of this paper. 


\subsection{Uncertainty Estimation}
Our work is also related to the uncertainty estimation, as uncertainty is crucial which allows an agent to discern when to exploit and when to explore its environment in RL~\cite{C2009Synthesis}. Previous intrinsic rewards can also be interpreted from the perspective of uncertainty estimation, which can evaluate curiosity by estimating the deep learning model's uncertainty (confidence). Take Disagreement as an example, instead of comparing the prediction to the ground-truth, they suggest to evaluate the uncertainty of multiple prediction models using the deep ensemble~\cite{ensemble}, despite incurring additional computation. RND also claims that the distillation error can be viewed as a quantification of the uncertainty. 
Unlike RND, in our work, we evaluate the uncertainty of given states though measuring the information gap between dual learners which rely on dynamic memory instead of a random fixed network.

\begin{algorithm}[tb]
 \caption{Dynamic Memory-based Curiosity}
 \label{alg1}
 \textbf{Initialization}: policy network $\pi_\phi$; dual online learner networks $f_{\theta_{1}}$, $f_{\theta_{2}}$; memory network $M_{\omega}$; coefficients of intrinsic and extrinsic reward $\zeta$, $\beta$.
 \begin{algorithmic}[1] 
  \WHILE{Training}
  \FOR {$t=1,\cdots,T$}
  \STATE Receive state $s_t$ from environment 
  \STATE $a_t \leftarrow \pi_\phi(a|s)$ based on policy network $\pi_\phi$
  \STATE Take action $a_t$, receive state $s_{t+1}$ and extrinsic reward ${r_{t}}^{ext}$ from environment
  \STATE Collect step data into replay buffer\cut{; store episodic data if an episode is finished} 
  \STATE $s_t\leftarrow s_{t+1}$
  \ENDFOR
  \STATE Sample batch data as $\left\{\left(s_{i}, a_{i}, {r_{i}}^{ext}, s_{i+1}\right)\right\}_{i=1}^{N}$from replay buffer
  \FOR {$each\,i = 1,\cdots,N$}
  \STATE Generate latent state vectors $z^{\theta_{1}}_{i} = f_{\theta_{1}}(s_i)$, $z^{\theta_{2}}_{i} = f_{\theta_{2}}(s_i)$, $z^{\omega}_{i} =M_{\omega}(s_i)$
  \STATE Calculate intrinsic reward $r^{int}_i = \|z^{\theta_{1}}_{i}-z^{\theta_{2}}_{i}\|^{2}$
  \STATE Calculate total reward ${r_{i}}^{total}=\zeta {r_{i}}^{int} + \beta r^{ext}_i$
  \ENDFOR
  
  \STATE  Update $\theta_{1}$ and $\theta_{2}$ with sampled data by minimizing loss with equation (\ref{update})
  \STATE Update $\omega$ with equation (\ref{omega})
  \STATE Update $\phi$ with sampled data by maximizing $r^{total}$ using RL algorithm
  \ENDWHILE
 \end{algorithmic}
\end{algorithm}

\begin{figure*}[t] 
    \centering
    \includegraphics[width=0.95\textwidth]{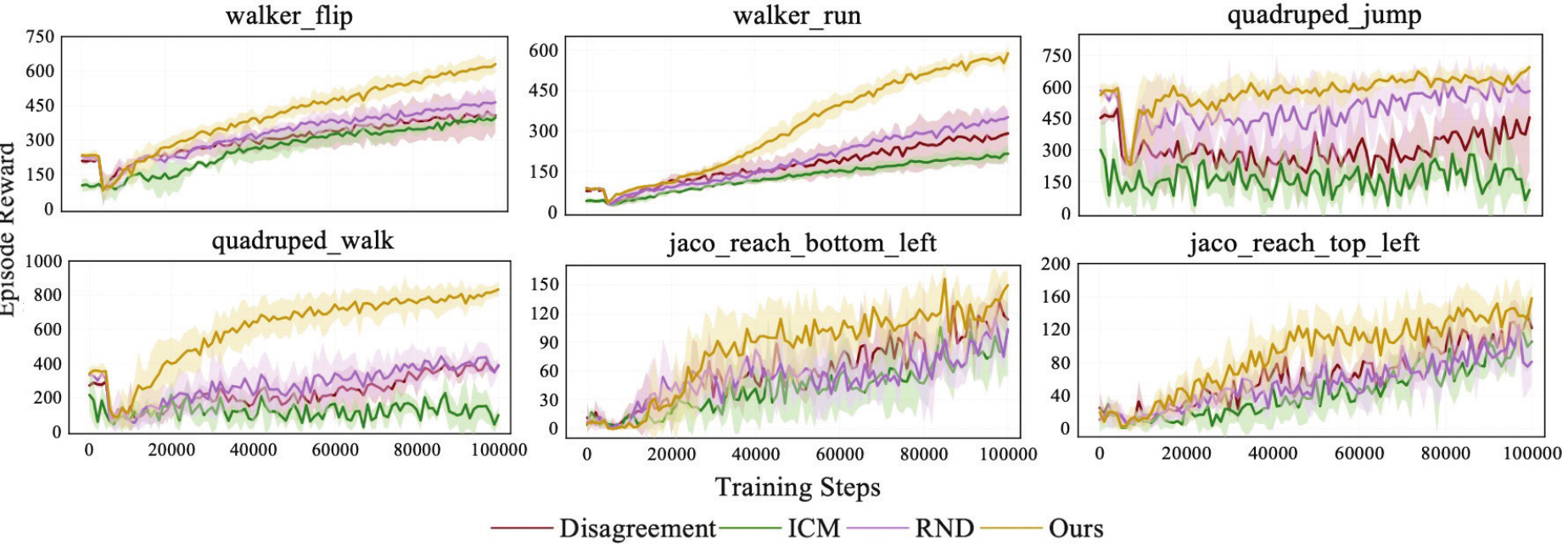}   
    \caption{\footnotesize{Performance of different methods in the fine-tuning phase on DeepMind Control Suite.}} 
    \label{fig-DMC-res}
\end{figure*}

\section{Methodology}

In general, if an agent encounters a state with the information value $E$ compared to its memory, then this state is worth exploring and such information value is worth consolidating to its memory dynamically~\cite{rotgans2017role,silvia2017curiosity}. In detail, the concept of information value $E$ necessitates the formation of the dynamic memory $M$ and a way $g$ to consolidate information to the memory. For deep neural networks, the memory $M$ can be embedded in the latent space and $g$ can by the function that maps state $s$ into memory~\cite{peterson2021curiosity}. \cut{Using the function-arrow notation, t}This kind of consolidating information is denoted by:
\begin{equation}
g(s;M) \rightarrow M^{\prime}.
\label{theory1} 
\end{equation}

With the memory $M$ which has been learned by $g$ over historical states, we can measure the information value $E$ of the next state $s_{t+1}$. According to the information theory~\cite{theory1,theory2,theory3} and the concepts proposed in \citet{peterson2021curiosity}, the information value $E$ of a state should (1) only depend on the memory and what can be immediately learned (i.e., $M$, $s$ and $g$); (2) be non-negative because $E$ is for exploring the environment; (3) decelerate in the finite time for the same state. Thus we define $E$ as:
\begin{equation}
E = \left\|g(s_{t+1};M)-M\right\|.
\label{theory2} 
\end{equation}

We can get from the definition of $E$ as :

(1) If one state has been completely explored, or cannot be learned, then no more information gain can be added into the current memory, and $E=0$. Such state is no longer worth exploring.

(2) If $E>0$, then the larger the value of $E$, the more information gain can be consolidated into the memory. In other words, the larger the value of $E$, the memory $M$ is less aware of the current state, such state is more worth exploring. It is such information deficiency of memory that sparks the curiosity of agents.

In this paper, we will focus on how to obtain and leverage the information value $E$ for agents exploration, and the information consolidation method $g$ in details.

\subsection{Dynamic Memory-based Curiosity}

In our framework, if the information in current memory cannot handle the encountered state, then the curiosity is aroused. We model the memory as a learnable neural network, but there is a dilemma that we do not have a ``benchmark'' encoded network in the parameter space to encode the encountered state accurately, since not enough supervised signal provided here. Thus it is difficult and not sound to directly define the curiosity by comparing a random encoded state with the output latent state from a dynamic memory network. We instead introduce dual online learners for state encode representations. These dual learners have the same network architecture as the memory network but with each own different parameters. In their network parameter space, the dual networks are supervised by the memory encoding ability. And then our curiosity can be defined by the gap of the encodings of the same state output by dual learners' networks. The intuition of our dual learners is: if the state information has been squeezed out by the memory, then the memory can completely know and resolve the state, and the dual learners which can be seen as the two imitators to the memory are easier to get the similar encodings to the current state. In other words, if one state is little known by the memory, then the dual learners may produce quite different encodings to it, which represents the larger information value $E$ and thus stimulates the agent to explore this state. Here, for the uniform description, we refer the encoded states and encodings to the latent states, which reflect the cognition of states by the memory and learner networks.

In our implementation, such kind of latent gap $E$ will spark the curiosity as the intrinsic reward for agents exploration. After the RL agents learning, such information will be consolidated to the memory for memory better growing. In terms of the consolidation method $g$, it is externalized as updating the memory parameters via the exponential moving average (EMA)~\cite{haynes2012exponential,klinker2011exponential,grebenkov2014following} of the dual learners' parameters.

From such analysis, we see the dual learners first learn based on the memory network for measure the information value for exploration, and then the memory network consolidate information gain based on dual learners in the EMA way. The memory is actually seeking for the appropriate position in the parameter space dynamically, in order that its network can better characterize the memory and cognition ability of the seen states in environments. In a word, our dynamic memory is updated in a bootstrap~\cite{bootstrap2} way.
Figure~\ref{method} and algorithm \ref{alg1} present the whole framework and pseudo-code of DyMeCu. 

\begin{table*}[t]
\centering
\caption{Performance comparison with different pre-training methods on DeepMind Control Suite. The best results are in bold font in each task, and the second best results are underlined.\cut{ numbers indicate optimal performance, whereas the underlined numbers indicate sub-optimal performance.}}
\label{table-DMC}
\resizebox{1.0\linewidth}{!}{    
\renewcommand{\arraystretch}{1.0} 
\begin{tabular}{cc|cccc|cccc|c}
\hline
Domain                      & Task               & ICM            & Disagreement    & RND             & APT    & ProtoRL & SMM    & DIAYN  & APS             & Ours                           \\ \hline
                            & Flip               & 398$\pm$18         & 407$\pm$75          & 465$\pm$62          & 477$\pm$16 & 480$\pm$23  &  \underline{505$\pm$26} & 381$\pm$17 & 461$\pm$24          & \textbf{630$\pm$37} \\
                            & Run                & 216$\pm$35         & 291$\pm$81 & 352$\pm$29          & 344$\pm$28 & 200$\pm$15  &  \underline{430$\pm$26} & 242$\pm$11 & 257$\pm$27          & \textbf{588$\pm$25}                                 \\
                            & Stand              &  \underline{928$\pm$18}         & 680$\pm$107          & 901$\pm$8           & 914$\pm$8  & 870$\pm$23  & 877$\pm$34 & 860$\pm$26 & 835$\pm$64          & \textbf{965$\pm$5}                       \\
\multirow{-4}{*}{Walker}    & Walk               & 696$\pm$162         & 595$\pm$153          & 814$\pm$116          & 759$\pm$35 & 777$\pm$33  & \underline{821$\pm$36} & 661$\pm$26 & 711$\pm$68          & \textbf{934$\pm$16}                        \\

\multicolumn{2}{c|}{\textit{Average Performance} }                                   & 560$\pm$59            & 494$\pm$104    & 633$\pm$54    & 624$\pm$22 & 582$\pm$24 &   \underline{659$\pm$31}   & 536$\pm$20  & 566$\pm$46             & \textbf{780$\pm$21}                           \\

\hline
                            & Jump               & 112$\pm$4         & 383$\pm$265          &  \underline{580$\pm$72}          & 462$\pm$48 & 425$\pm$63  & 298$\pm$39 & 578$\pm$46 & 529$\pm$42          & \textbf{694$\pm$15}                        \\
                            & Run                & 91$\pm$29         & 389$\pm$61          & 385$\pm$47          & 339$\pm$40 & 316$\pm$36  & 220$\pm$37 &  \underline{415$\pm$28} & 465$\pm$37 & \textbf{479$\pm$6}                                 \\
                            & Stand              & 184$\pm$100         & 628$\pm$114          & 800$\pm$54          & 622$\pm$57 & 560$\pm$71  & 367$\pm$42 & 706$\pm$48 &  \underline{714$\pm$50}          & \textbf{921$\pm$14}                                 \\
\multirow{-4}{*}{Quadruped} & Walk               & 99$\pm$46         & 384$\pm$28          & 392$\pm$39 & 434$\pm$64 & 403$\pm$91  & 184$\pm$26 & 406$\pm$64 &  \underline{602$\pm$86}          & \textbf{833$\pm$44}                               \\
\multicolumn{2}{c|}{\textit{Average Performance} }                                   & 122$\pm$45            & 446$\pm$117    & 540$\pm$53    & 465$\pm$52    & 426$\pm$66 &  268$\pm$36   & 527$\pm$47  &  \underline{578$\pm$43}             & \textbf{732$\pm$20}                           \\
\hline
                            & Reach bottom left  & 102$\pm$47 & 117$\pm$17           & 103$\pm$17          & 88$\pm$12  &  \underline{121$\pm$22}  & 40$\pm$9   & 17$\pm$5   & 96$\pm$13           & \textbf{155$\pm$13}                        \\
                            & Reach bottom right & 75$\pm$27          &  \underline{142$\pm$3}           & 101$\pm$26          & 115$\pm$12 & 113$\pm$16  & 50$\pm$9   & 31$\pm$4   & 93$\pm$9            & \textbf{146$\pm$16}                      \\
                            & Reach top left     & 105$\pm$29         & 121$\pm$17          &  \underline{146$\pm$46}          & 112$\pm$11 & 124$\pm$20  & 50$\pm$7   & 11$\pm$3   & 65$\pm$10           & \textbf{166$\pm$14}                       \\
\multirow{-4}{*}{Jaco}      & Reach top right    & 93$\pm$19          & 131$\pm$10           & 99$\pm$25          & \underline{136$\pm$5}  & 135$\pm$19  & 37$\pm$8   & 19$\pm$4   & 81$\pm$11           & \textbf{152$\pm$4}        \\
\multicolumn{2}{c|}{\textit{Average Performance} }                                   & 94$\pm$31            &  \underline{128$\pm$12}    & 113$\pm$29   & 113$\pm$10    & 124$\pm$20 &  44$\pm$9   & 20$\pm$4  & 84$\pm$11             & \textbf{155$\pm$12}                           \\
\hline  
\end{tabular}}
\end{table*}
\begin{itemize}
\item \textbf{Learning of Dual Learners:}
\end{itemize}

Dual online learner models $f_{\theta_1}$ and $f_{\theta_2}$ are defined by a set of weights $\theta_1$ and $\theta_2$ with the same architecture as the memory network $M_{\omega}$. The memory provides the regression targets for the learning of dual learners $f_{\theta_1}$ and $f_{\theta_2}$. Given a current state $s_t$, the learners transform it into the latent states $z_{t}^{\theta_1} \triangleq f_{\theta_1}\left(s_{t}\right)$ and $z_{t}^{\theta_2} \triangleq f_{\theta_2}\left(s_{t}\right)$ respectively, and the memory network outputs $z_{t}^{\omega}\triangleq M_{\omega}\left(s_{t}\right)$. The mean squared error (MSE) between them is:
\begin{equation}
\left\{
\begin{aligned}
\mathcal{L}_{\theta_1} &\triangleq\left\|z_{t}^{\theta_1}-z_{t}^{\omega}\right\|^{2},\\
\mathcal{L}_{\theta_2} &\triangleq\left\|z_{t}^{\theta_2}-z_{t}^{\omega}\right\|^{2}.
\end{aligned}
\right.
\label{update}
\end{equation}

Based on $\mathcal{L}_{\theta_1}$ and $\mathcal{L}_{\theta_2}$, the dual learners are updated as :

\begin{equation}
\left\{\begin{aligned}
\theta_{1} &\leftarrow \operatorname{optim}\left(\theta_{1}, \nabla_{\theta_{1}} \mathcal{L}_{\theta_{1}}, \eta\right), \\ 
\theta_{2} &\leftarrow \operatorname{optim}\left(\theta_{2}, \nabla_{\theta_{2}} \mathcal{L}_{\theta_{1}}, \eta\right), \\ 
\end{aligned}\right.
\end{equation}
where $\operatorname{optim}$ and $\eta$ represent the optimizer and learning rate.

\begin{itemize}
\item \textbf{Intrinsic Reward based on Curiosity:}
\end{itemize} 

The curiosity relies on the information value of current state. In our method, such information value can be measured by the information gap between dual learners. This information gap can also be considered following the $\delta$-Progress ~\cite{progress1,progress2} to form the curiosity. We obtain the intrinsic reward to agents based on the curiosity from the information value:

\begin{equation}
r_{t}^{int}=\|(z_{t}^{\theta_1}-z_{t}^{\omega})-(z_{t}^{\theta_2}-z_{t}^{\omega})\|^{2}=\|z_{t}^{\theta_1}-z_{t}^{\theta_2}\|^{2}.
\label{eq4} 
\end{equation}

From another point of view, the dual-learner mechanism can be regarded as the variant of ensemble~\cite{mai2022sample} for uncertainty estimation. Compared with previous attempts which requires heavily ensembling (such as the Disagreement), our lightweight solution can previous better performance while retaining computation efficiency. 

Overall, we can get the optimization goal for the agent:
\begin{equation}
    \max_{\phi}~ \mathbb{E}_{\pi_{\phi}(s_t)}\left[\sum \gamma^{t} (\zeta r_{t}^{int} + \beta r_{t}^{ext})\right],
\end{equation}
where $\gamma$ is the discount factor and $\phi$ represents parameters of policy $\pi$; $\zeta$ and $\beta$ are the coefficients of the intrinsic reward and extrinsic reward respectively. 

\begin{itemize}
\item \textbf{Consolidating Information into Memory:}
\end{itemize}

The memory model is updated in an EMA way for sake of its stability to the old state information and the plasticity to the current new state information. In other words, the memory is dynamically growing taking the contextual environment information into account. Specifically, given a decay rate $\alpha \in[0,1]$ and after each training step, the memory $M_\omega$ can be updated as:
\begin{equation}
\omega \leftarrow \alpha \omega+(1-\alpha)\frac{{\theta_1}+{\theta_2}}{2}.
\label{omega}
\end{equation}

\cut{
\begin{itemize}
\item \textbf{Intuitions on DyMeCu’s behavior:}
\end{itemize} 
In this paper, we aim to form the dynamic memory to evaluate the curiosity instead of a random fixed target network, which is beneficial for dynamic assessment of curiosity and closer to human curiosity. From the view of semi-supervised learning, the memory model can also be regarded as the teacher model in Mean Teacher-based approach~\cite{mean} which is well-known as a knowledge distillation method. The memory is essentially a self-ensemble of the intermediate models of learners. The paradigm we proposed is one type of replay mechanism that is thought to play an important role in memory formation, retrieval, and consolidation~\cite{replay}. Similarly, the way we form the memory is also used in continual learning to address the issue of catastrophic forgetting~\cite{fast-slow}. In our opinion, measuring the curiosity for a given state is highly related to the measurement of how well the state is consolidated in memory. The more consolidated the current state is in memory, the less information value is. Thus, the curiosity can be smaller. So, we design the intrinsic reward as shown in~\ref{eq4} using the dual learner networks to evaluate the information value. The architecture and hyper-parameters of our networks are shown in Appendix.
}

\begin{itemize}
\item \textbf{Intuitions on DyMeCu’s behavior:}
\end{itemize} 

The dynamic memory-based curiosity is closer to human curiosity mechanism. It is the cognitive difference compared to the memory that stimulates our curiosity to explore the world, and then we will consolidate the cognition information to the memory dynamically. In addition, from the knowledge distillation view, such memory can also be regarded as the teacher model in Mean Teacher-based approach~\cite{mean}. The memory is essentially a self-ensemble of the intermediate models of learners. The paradigm we proposed is one type of replay mechanism that is thought to play an important role in memory formation, retrieval, and consolidation~\cite{replay}. Moreover, we consider our way to form the memory can also be used in the continual learning to address the issue of catastrophic forgetting~\cite{fast-slow}. 

 \begin{figure*}[t] 
    \centering
    \includegraphics[width=0.95\textwidth]{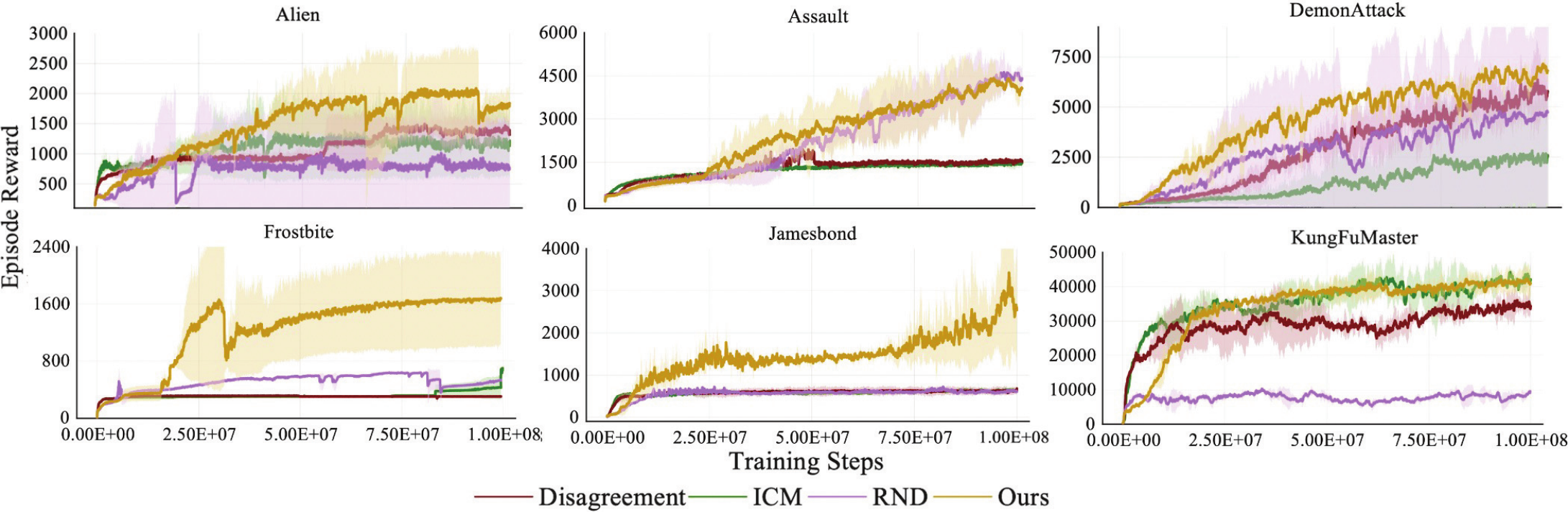}   
    \caption{Performance comparison on Atari games subsets using both intrinsic rewards and extrinsic rewards.} 
    \label{int+ext_6}
\end{figure*}

\section{Experiments and Analysis}

\subsection{Experimental Settings}

\cut{To conduct quantitative and comprehensive comparisons, w}
We evaluate our method in both pre-training and traditional RL situations utilizing two widely used benchmarks: DeepMind Control Suite (DMC)~\cite{dm_control} and Atari Suite~\cite{atari}.
We follow the RND~\cite{RND} experimental settings for Atari Suite and settings of URLB~\cite{urlb} which is the pre-training benchmark for DeepMind Control Suite. We apply PPO algorithm~\cite{schulman2017proximal} to train the agent.
The hyper-parameter $\alpha$ was set as 0.99 in all experiments, and\cut{ all comparison experiments are based on the same settings. T} the implementation details and hyper-parameters can be found in the appendix.

\subsection{DeepMind Control Suite}
Many well-performing approaches like URLB~\cite{urlb} use the pre-training and fine-tuning paradigm to improve sample efficiency for RL, especially in the experiment benchmark like DMC containing various domains and complex tasks.
We evaluate DyMeCu on all three domains of DMC, namely Walker, Quadruped, and Jaco Arm (from easiest to hardest), and each of them has four tasks. During the pre-training phase, the agents are trained for 2 million steps with only intrinsic rewards produced by the curiosity. During the fine-tuning phase, the agents are trained for 100k steps with only extrinsic rewards.

Table~\ref{table-DMC} reports the final scores and standard deviations of DyMeCu and other competitive methods. We compare DyMeCu with both intrinsic reward-based methods (ICM, Disagreement, RND, APT~\cite{liu2021behavior}) and other pre-training strategies (ProtoRL~\cite{prot}, SMM~\cite{lee2019efficient}, DIAYN~\cite{eysenbach2018diversity}, APS~\cite{liu2021aps}). DyMeCu improves average performance by 18.3\%, 26.6\%, and 21.0\% on these three domains respectively. From the quantitative results, we can see our DyMeCu achieve the new state-of-the-art across all 12 tasks, demonstrating DyMeCu's ability to improve the model performance and robustness through pre-training paradigm.
Figure~\ref{fig-DMC-res} plots 6 learning curves (fine-tuning phase) of DyMeCu and three competitive curiosity-based methods. All learning curves can be found in the appendix. 
DyMeCu shows a superior convergence speed than other methods. Meanwhile, the convergence result of DyMeCu also surpasses others significantly. DyMeCu's speed increase may sbe mainly due to the contextual state information being consolidated into memory dynamically, rather than a random fixed setting like the RND. Based on the dynamic memory, the exploration of agents can be much more efficient.

\subsection{Atari Suite}
For the Atari suite, we first record the performance of agents with both intrinsic and extrinsic rewards. The experiments conducts 100M running steps - equivalent to 400M frames and the intrinsic and extrinsic rewards coefficients are set to $\zeta=1$ and $\beta=2$ respectively for all methods, following the setup of the previous curiosity-based methods.  Table~\ref{table-int+ext} lists the aggregate metrics and scores of three methods trained with both intrinsic and extrinsic rewards on the Atari 26 games. Human and random scores are adopted from~\citet{rainbow}. As done in previous works \cite{liu2021behavior,yarats2020image,SPR2020data}, we normalize the episode reward as human-normalized scores (HNS) by expert human scores to account for different score scales in each game. \#SOTA denotes the number of games that the current method exceeds other methods and mean HNS is calculated as the average of $(\text {agent score}-\text {random score})/(\text {human score}-\text {random score})$ of all games. From Table~\ref{table-int+ext}, DyMeCu displays the superiority over Disagreement and ICM with its highest mean HNS and \#SOTA.
\cut{
, since its mean HNS surpasses both baselines and the game numbers of exceeding other approaches is also greater than other baselines.
}
\begin{table}[t]
\footnotesize
\caption{Performance comparison of curiosity-based methods using both intrinsic and extrinsic rewards on 26 Atari games subset. The bold font indicates the best value.}
\centering
\label{table-int+ext}
\resizebox{1.0\linewidth}{!}{    
\renewcommand{\arraystretch}{1.0} 
\begin{tabular}{lll|lllll}
\hline
Game           & Random   & Human   & ICM     & Disagreement  & Ours \\ \hline
Alien          & 227.8    & 7127.7  & 1524.7  & 1304.7        & \textbf{2589.2} \\
Amidar         & 5.8      & 1719.5  & \textbf{763.0}   & 506.6       &   470.1    \\
Assault        & 222.4    & 742.0   & 1365.5  & 1544.6        & \textbf{4539.3} \\
Asterix        & 210.0    & 8503.3  & 2103.4  & 1616.2       & \textbf{4576}\\
Bank Heist     & 14.2     & 753.1   & 1359.4  & 1343.4       &    \textbf{1529.5}    \\
BattleZone     & 2360.0   & 37187.5 & 51459.1 &   \textbf{65387.4}         &    58220.0    \\
Boxing         & 0.1      & 12.1    & 98.9    &      99.3       &    \textbf{99.6}   \\
Breakout       & 1.7      & 30.5    & \textbf{247.6}   &     177.8       &   119.7   \\
ChopperCommand & 811.0    & 7387.8  & 9456.5  &    \textbf{10286.9}         &   9521.0    \\
Crazy Climber  & 10780.5  & 23829.4 & \textbf{135003.3}    &    132614.0     &    106682.0 \\
Demon Attack   & 107805.0 & 35829.4 &   4679.2      &       6606.0      &    \textbf{8417.0}   \\
Freeway        & 0.0      & 29.6    &     33.8    &     \textbf{33.9}        &    30.7   \\
Frostbite      & 65.2     & 4334.7  &    309.4     &    295.1        &    \textbf{1750.0}   \\
Gopher         & 257.6    & 2412.5  &    \textbf{14619.4}     &    14202.4    &   9750.3    \\
Hero           & 1027.0   & 30826.4 &  13482.2      &      \textbf{13488.0}     &    12728.5   \\
Jamesbond      & 29.0     & 302.8   & 680.8   & 726.5    & \textbf{5052.5} \\
Kangaroo       & 52.0     & 3035.0  &    12922.7     &      \textbf{14621.8}   &   10760.0   \\
Krull          & 1598.0   & 2665.6  &   10027.1      &      \textbf{11402.7}     &  6447.0     \\
Kung Fu Master & 258.5    & 22736.3 &    40157.7     &      32607.2     &  \textbf{44604.9}     \\
Ms Pacman      & 307.3    & 6951.6  & 2787.0  & \textbf{6287.8}   & 2752.4 \\
Pong           & -20.7    & 14.6    &  20.1       &     \textbf{20.6}      &   15.3     \\
Private Eye    & 24.9     & 69571.3 &    96.0     &    98.0    &   \textbf{100.0}     \\
Qbert          & 163.9    & 13455.0 &    16388.9     &      \textbf{22474.5}    &    14770.2    \\
Road Runner    & 11.5     & 7845.0  &      \textbf{56273.7}   &      55359.3  &    32271.0    \\
Seaquest       & 68.4     & 42054.7 &   \textbf{16178.1}      &      2733.1      &    3910.9    \\
Up N Down      & 533.4    & 11693.2 & \textbf{46152.9} & 18235.5     & 18067.6 \\ \hline
Mean HNS       & 0.0      & 1.0     &   2.861      &       2.767    &  \textbf{3.019}     \\
\#SOTA       & N/A      & N/A     &     7    &          9   &  \textbf{ 10 }   \\ 

\hline
\end{tabular}
}
\end{table}

\begin{figure*}[t] 
    \centering
    \renewcommand{\thefigure}{4}
    \includegraphics[width=0.95\textwidth]{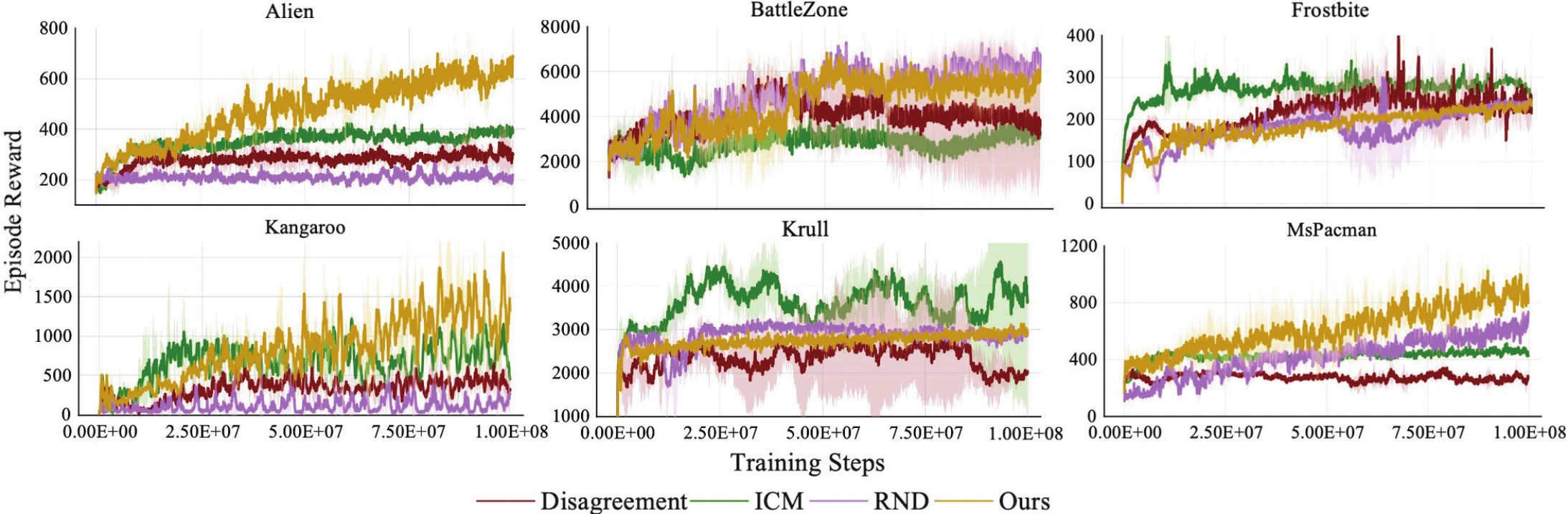} 
    \caption{Performance comparison on Atari games suite subsets using only intrinsic rewards.} 
    \label{int6}
\end{figure*}

Figure~\ref{int+ext_6} displays the learning curves using both intrinsic and extrinsic rewards. We compare DyMeCu with three widely-used baselines, including  Disagreement, ICM and RND, on 6 random chosen Atari games. DyMeCu shows evident advantages in most games on the performance and learning speed. For example, on Jamesbond, the convergence plot reward of DyMeCu is more than three times that of other methods.
Moreover, we also compare the performance of agents trained with only intrinsic rewards. As shown in Figure~\ref{int6}, of the 6 environments, DyMeCu outperforms Disagreement baseline in all environments, outperforms ICM and RND baselines both in 4 environments.\cut{ our method has also got minor variance compared with other methods, proving the effectiveness of DyMeCu further.}\cut{ We can find the similar phenomenon in human agents, where individual behavioral boundaries are more predictable with the contextual information.}
Overall, the results in Atari Suite show that DyMeCu outperforms other curiosity-based methods, demonstrating DyMeCu's ability to generate more accurate intrinsic rewards and provide more useful information for better exploration.

\cut{\subsection{Building intuitions with ablations}}
\subsection{Further Analysis on DyMeCu}
Further analysis including ablation studies on DyMeCu are presented to give an intuition of its behavior and performance. \cut{For reproducibility, we execute each parameter setting across three seeds and provide the average performance. }We run the experiments across 3 random seeds and all following experiments conducts 50M running steps - equivalent to 200M frames.
\begin{figure}[t] 
    \centering
    \includegraphics[width=1.0\linewidth,scale=1.0]{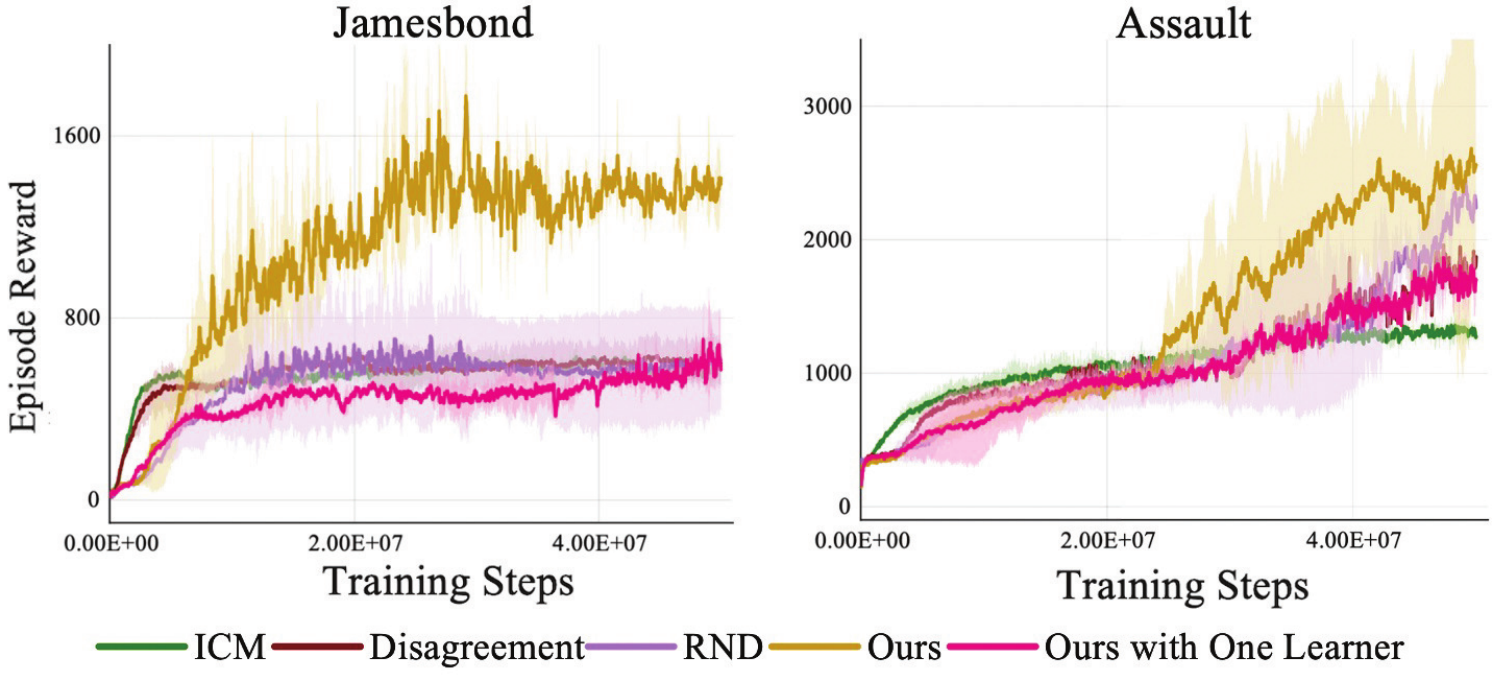}
    \renewcommand{\thefigure}{5}
    \caption{Performance comparison among two kinds of deployments and baselines with both intrinsic and extrinsic rewards.}
    \label{One learner}
\end{figure}

\begin{itemize}
\item \textbf{Dual learners:}
\end{itemize}

Here we explore to design the curiosity under the naive setting, that is, using one encoding network to learn to encode the latent space, and thus the curiosity-based intrinsic reward can be defined as the gap with the memory network:

\cut{We explored to design the curiosity using only one online learner. In the one-learner settings, the intrinsic reward can be defined as follows:}
\begin{equation}
r_{t}^{int}=\|z_{t}^{\theta}-z_{t}^{\omega}\|^{2}.
\label{eq5} 
\end{equation}

The memory is updated with $\omega \leftarrow \alpha \omega+(1-\alpha){\theta}$. 
As shown in Figure~\ref{One learner}, one-learner mechanism does not show significant advantages over other methods, whereas our dual-learner mechanism performs much better with more accurate curiosity and corresponding intrinsic rewards.

\begin{table}[t]
\footnotesize
\caption{Performance comparison of baselines and DyMeCu under different settings with only intrinsic rewards. The results represent the average episode reward at the end of training. The Ave. in the last column shows the average result among the three tasks.}
\centering
\label{table-ablation}
\resizebox{1.0\linewidth}{!}{    
\renewcommand{\arraystretch}{1.0}
\begin{tabular}{c|c|c|c|c}
\hline \diagbox{Method}{Game} & Alien & Kangaroo & MsPacman &Ave. \\
\hline Disagreement & 316.6 & 514.0 & 291.0 &373.9 \\
\hline ICM & 374.2 & 557.0 & 412.7 & 447.9 \\
\hline RND & 206.1 & 412.0 & 607.2 & 408.4 \\
\hline DyMeCu (ours) & 492.0 & 739.0 & 602.4 & 611.1 \\
\hline DyMeCu\_{update with one learner}  & 521.7 & 782.0 & 500.6 & 601.4 \\
\hline DyMeCu\_{with additional module} & 390.6 & 645.2 & 644.4 & 560.1\\
\hline
\end{tabular}
}
\end{table}

\begin{itemize}
\item \textbf{Update of memory network:}
\end{itemize}

The memory network in DyMeCu is updated with dual learners, we additionally evaluate the performance of DyMeCu when the memory is updated using only one of the learner's parameters.
The results in Table~\ref{table-ablation} indicate that both learners can consolidate state information into the memory well. Combined with Figure~\ref{One learner}, it is useful and necessary to assign and train dual learners, and then we can update the memory with dual or one-learner, while dual-learner update mechanism shows a little superior performance.

\begin{itemize}
\item \textbf{Structure of learners:}
\end{itemize}
The bootstrap idea has been explored and used in some previous researches. The most similar one to ours is BYOL~\cite{bootstrap2}, which uses the bootstrap method for self-supervised learning in computer vision. Furthermore, \citeauthor{bootstrap2} add another predictor module to the online network, and compare the output of predictor to the target network, and it is the key to generating well-performed representations~\cite{sia}. 
Similarly, in this ablation study, we also design the controlled trials, in which additional two convolution layers are added to each of dual learners. 
In Table~\ref{table-ablation}, we can find that such learnable additional module does not lead to significant improvement. Under our analysis, unlike previous work using the bootstrap method, we aim to generate the intrinsic reward by calculating the information value (i.e., information gap between dual learners) as accurate as possible, instead of better representations for downstream tasks.

\begin{itemize}
\item \textbf{Robustness to hyper-parameter $\alpha$:}
\end{itemize}
\begin{figure}[t] 
    \centering
    \includegraphics[width=0.8\linewidth,scale=1.0]{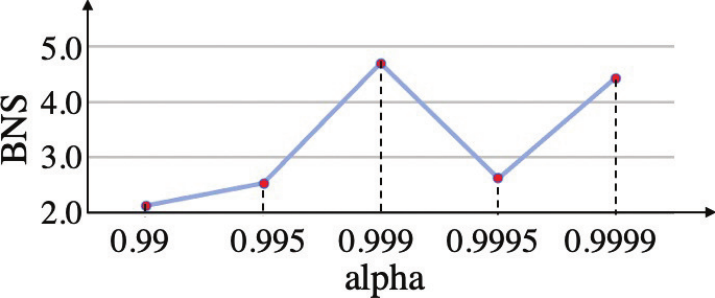}   
    \caption{Performance comparison with different values of $\alpha$ with only intrinsic rewards.}
    \label{alpha}
\end{figure}

There is a concern of the updating speed of memory network in the EMA way. It is about how much and how fast to accept and consolidate the new environment information.
Therefore, to further analyze the updating effect of the hyper-parameter $\alpha$, we evaluate DyMeCu with different values of $\alpha$ in a rational interval, and we assess the agents' performance in three different Atari games: Alien, Kangaroo, and Krull. \cut{To compare with baselines more visually}For more direct and visual comparison, we normalize the episode reward of DyMeCu as baseline-normalized scores (BNS) which is calculated as the average of 
$(\text {DyMeCu score}-\text {random score})/(\text{baseline score}-\text{random score})$ where the $\text{baseline score}$ is the average score of baselines. As illustrated in Figure~\ref{alpha}, all values of the hyper-parameter $\alpha$ between 0.99 and 0.9999 yield satisfied performance, generally greater than twice that of the baseline average. DyMeCu shows acceptable robustness to the updating hyper-parameter.

\section{Conclusion}
To address the challenge of extrinsic rewards sparsity in RL, we propose DyMeCu to mimic human curiosity in this paper. Specifically, DyMeCu consists of a dynamic memory and dual online learners. The information gap between dual learners sparks the agent's curiosity and then formulates the intrinsic reward, and the state information can then be consolidated into the dynamic memory. Large-scale empirical experiments are conducted on multiple benchmarks, and the experimental results show that DyMeCu outperforms competing curiosity-based methods under different settings.

\bibliography{aaai23}
\end{document}